\documentclass[conference]{IEEEtran}
\IEEEoverridecommandlockouts
\usepackage{cite}
\usepackage{amsmath,amssymb,amsfonts}
\usepackage{algorithmic}
\usepackage{graphicx}
\usepackage{textcomp}
\usepackage{xcolor}
\usepackage{subcaption}
\usepackage{array}
\def\BibTeX{{\rm B\kern-.05em{\sc i\kern-.025em b}\kern-.08em
    T\kern-.1667em\lower.7ex\hbox{E}\kern-.125emX}}
    
\newcolumntype{P}[1]{>{\centering\arraybackslash}p{#1}}
    
\begin{document}

\title{Genome Sequence Classification for Animal Diagnostics with Graph Representations and Deep Neural Networks\\
\thanks{This project is funded by US Department of Agriculture (USDA)}
}

\author{\IEEEauthorblockN{Sai Narayanan\IEEEauthorrefmark{2}, Akhilesh Ramachandran\IEEEauthorrefmark{2}, Sathyanarayanan N. Aakur\IEEEauthorrefmark{1}, Arunkumar Bagavathi\IEEEauthorrefmark{1}}
\IEEEauthorblockA{\textit{Dept. of Computer Science\IEEEauthorrefmark{1}, Dept. of Veterinary Medicine\IEEEauthorrefmark{2}} \\
\textit{Oklahoma State University, Stillwater, OK, 74078}\\
ssankar@okstate.edu, rakhile@okstate.edu, saakurn@okstate.edu, abagava@okstate.edu}
}

\maketitle

\begin{abstract}
\textit{Bovine Respiratory Disease Complex} (BRDC) is a complex respiratory disease in cattle with multiple etiologies, including bacterial and viral. It is estimated that mortality, morbidity, therapy, and quarantine resulting from BRDC account for significant losses in the cattle industry. Early detection and management of BRDC are crucial in mitigating economic losses. Current animal disease diagnostics is based on traditional tests such as bacterial culture, serolog, and Polymerase Chain Reaction (PCR) tests. Even though these tests are validated for several diseases, their main challenge is their limited ability to detect the presence of multiple pathogens simultaneously. Advancements of data analytics and machine learning and applications over metagenome sequencing are setting trends on several applications. In this work, we demonstrate a machine learning approach to identify pathogen signatures present in bovine metagenome sequences using k-\textit{mer}-based network embedding followed by a deep learning-based classification task. With experiments conducted on two different simulated datasets, we show that networks-based machine learning approaches can detect pathogen signature with up to $89.7\%$ accuracy. We will make the data available publicly upon request to tackle this important problem in a difficult domain. 
\end{abstract}


\section{Introduction}
Pathogen identification and disease diagnostics is an ever-evolving field in veterinary medicine. The earliest and accurate pathogen detection has always been the goal of infectious disease diagnosis. Traditional methods such as bacterial and viral cultures, even though considered gold standard tests, require days to weeks to identify pathogens. Testing protocols, like Polymerase Chain Reaction (PCR) tests, based on the detection of unique DNA/RNA elements, are faster and considered to be very accurate for most of the infectious diseases in animals~\cite{maurin2012real} and human beings~\cite{elnifro2000multiplex}. Although well established, PCR imposes several challenges. Conventional PCR protocols are target-specific with limited multiplexing capability. This requires the need to design and develop pathogen-specific probes and primers that can function under selected thermocyclic parameters to target multiple pathogens. Unless specifically targeted, these methods provide very little information regarding other co-infections or innate susceptibilities based on the genetic make-up of the host or alterations in commensal microbiomes. This necessitates the development and maintenance of multiple target-specific standard operating protocols (SOPs), which can be significantly time-consuming for National Animal Health Laboratory Network (NAHLN) laboratories developing these protocols and the member laboratories participating in testing services.

Genetic sequencing-based techniques have proven importance in human~\cite{koser2012routine} and veterinary diagnostics~\cite{thomsen2016bacterial}. Decreasing trends in hardware and testing costs seen over recent years make these techniques affordable for diagnostic applications. Specifically, shotgun metagenome sequencing aims to sequence total genetic material from all sources in a clinical sample (i.e., from the host, pathogens, commensals, environmental components, etc.) without introducing bias~\cite{sharpton2014introduction}. Advancements in data analytics and machine learning have a significant impact on genome sequencing over multiple domains. 
Developing machine learning models to analyze large volumes of metagenome sequence data rapidly can immensely advance the field of animal disease diagnostics and help in the detection of known and emerging pathogens in a single test.

\begin{figure*}
    \centering
    \includegraphics[scale=0.7]{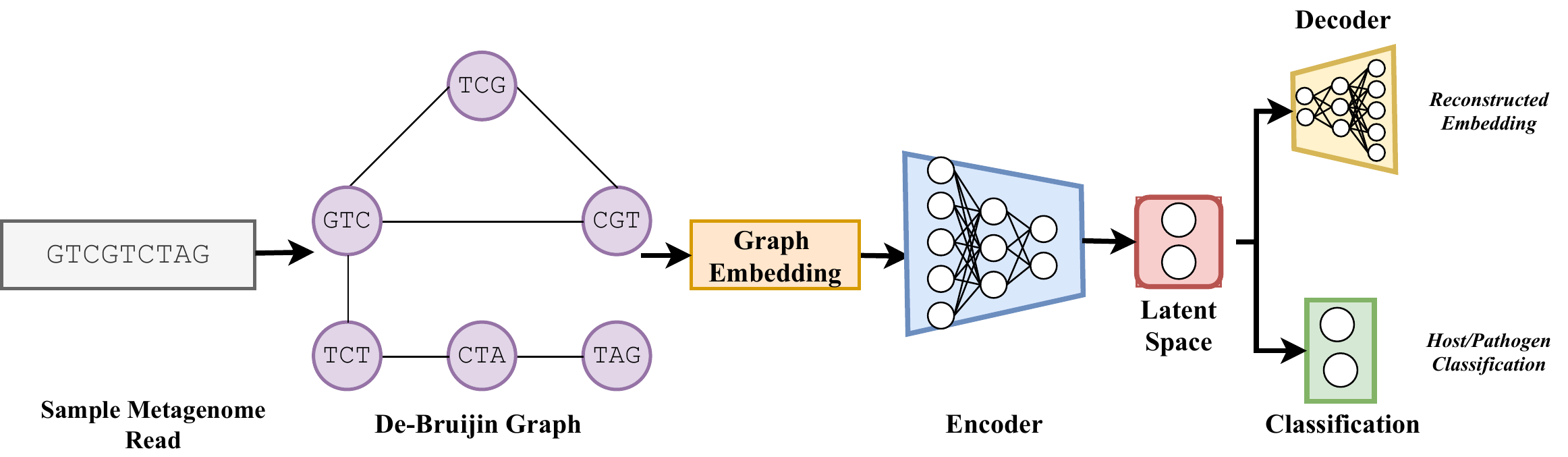}
    \caption{The overall framework for the proposed metagenome classification pipeline.}
    \label{fig:framework}
\end{figure*}

Large scale graphs/networks based machine learning tasks are continuously proving their importance in several domains like social science~\cite{bagavathi2019examining}, biochemistry~\cite{zhang2017network}, and biology~\cite{barabasi2020genetic}. One of such crucial tasks is to extract \emph{node} and \emph{graph} features, commonly known as \emph{network embedding} from their complex and unstructured organization in graph topologies. Network embedding can be extracted in multiple ways ranging from \emph{graph kernels} to \emph{feature extraction}. Graphs or graph vertexes mapped to low dimensional feature vector space are utilized for machine learning applications like classification and clustering. Deep neural networks have shown a tremendous capacity to capture complex patterns in high dimensional data, particularly in images~\cite{krizhevsky2012imagenet}, text~\cite{devlin2018bert}, and even hardware security~\cite{ramnath2019latent}. There have been emerging applications to learn robust representations from structured data such as graphs~\cite{scarselli2008graph,fout2017protein}. We aim to build on these advances in a unified framework for genome sequence classification for animal diagnostics, particularly for detecting the single bacterial (\textit{Mannheimia hemolytica}) infection called \textit{Bovine Respiratory Disease Complex} (BRDC) in bovine (cattle) metagenomes. It is particularly important to identify this infection since it migrates to the lower respiratory tract and can compromise the immunity of the infected animal(as pneumonia) and can spread rapidly among crowded groups~\cite{klima2014characterization}. Hence, the diagnosis of BRDC at the earliest plays a major role in mitigating losses.

In this work we represent genome sequences as \emph{De-Bruijn graphs}, experiment multiple network representation learning techniques including multiple graph kernels~\cite{siglidis2020grakel}, node2vec~\cite{grover2016node2vec}, and graph2vec~\cite{narayanan2017graph2vec} to obtain representational features from constructed \emph{De-Bruijn graphs}, and use classic machine learning and deep learning methods to identify pathogens in animal genome sequences. In particular, we use the Bovine Respiratory Disease Complex (BRDC) to validate the use of machine learning-based animal diagnostics. In this initial phase of our research, we model the single bacterial pathogen detection problem as a classification task with existing graph-based machine learning approaches. 
%
Our overall approach is shown in Figure~\ref{fig:framework}. We construct De-Bruijn graphs (Section~\ref{sect:kmer} to create a standard, structured representation of genome sequences. We then build vector-representations of the graph through various, existing embedding models (Section~\ref{sec:embed}). 
The graph embedding is then used to distinguish between pathogen and host genome sequences using a deep neural network (Section~\ref{sec:method}), trained in a multi-task learning~\cite{caruana1997multitask} setting. 

Our \textbf{contributions} are three-fold: (i) we present the first large-scale, annotated genome sequence dataset for the diagnosis of a bacterial pathogen from bovine metagenome sequences, (ii) show that De-Bruijn graphs can be extended to the diagnosis task using network embedding and deep neural networks, and (iii) to the best of our knowledge, our work obtains the state-of-the-art in utilizing graph-based representations to classify pathogens from very large metagenome sequences.

\section{Related Work}

\subsection{Genome Sequencing with Machine Learning}
Diagnosis of pathogen genome sequences within larger animal metagenomes have traditionally been tackled through bioinformatics approaches such as k-\textit{mer} frequency-based features~\cite{koren2017canu}. The k-\textit{mer} representation is a compositional feature representation in bioinformatics that capture the frequency of the presence of k-length subsequences within a larger genome sequence. Such representations have shown to be effective in several metagenome diagnostic tasks such as chromatin accessibility prediction~\cite{min2017chromatin} and bacteria detection~\cite{deneke2017paprbag,fiannaca2018deep}, but have not hard much success in longer sequence reads. Deep learning and machine learning models have provided an automatic, higher-order feature learning beyond pre-defined motif lengths defined in k-\textit{mers}. Convolutional neural networks (CNNs), primarily used in computer vision~\cite{krizhevsky2012imagenet}, have shown great success in genome sequence prediction and classification by capturing powerful, hierarchical feature representations~\cite{nguyen2017deep}. Sequence-based models such as Recurrent Neural Networks and Long Short-Term Memory (LSTM) networks~\cite{hochreiter1997long} have also been implemented to capture long-term dependencies in genome sequences~\cite{min2017chromatin} successfully.

\subsection{Genome Sequencing with Networks}
Graphs or Networks have been a widely used structure to study genomics for a variety of problems. Examples include, genomic sub-compartment prediction~\cite{ashoor2020graph}, disease gene predictions~\cite{hwang2019humannet}, genome assembly~\cite{lin2016assembly} to name a few. The most common approach to represent genome sequences in networks is with k-\textit{mers}, where a long genome sequence is broken into \emph{k-pairs} of shorter sketches~\cite{marccais2018asymptotically}. Over the years, the bioinformatics research community has introduced a variety of graph structures, using k-\textit{mers}, to study genome sequences. Some of them include: \emph{De Bruijn graphs}~\cite{pevzner2001eulerian} and \emph{STRING graphs}~\cite{myers2005fragment}, which merges repeating genome patterns into one node in the network, \emph{linked De Bruijn graphs}~\cite{turner2018integrating}, which include metadata of genome sequences to store connectivity information, and variational De Bruijn graphs like \emph{pufferfish}~\cite{almodaresi2018space}, which is introduced for efficient query processing.


\section{Preliminaries}

\begin{table}[h!]
	\centering
	
	\caption{Datasets Summary and Basic Statistics of De-Bruijn Graphs used in Experiments}
	\begin{tabular}{p{10em}|p{6.5em}|p{6.5em}}
		\hline

		\textbf{Dataset} & \textbf{DS500 \textit{k=6}} & \textbf{DS5000 \textit{k=3}} \\
		\hline
	
		\textbf{\# of positives/negative samples} & 500 & 5000\\
		\hline
		
		\textbf{Total samples/graphs} & 1000 & 10000\\
		\hline
		
		\textbf{k} & 6 & 3\\
		\hline
		\hline
	
		\textbf{Avg. no. of nodes} & $139 \pm 4.84$ & $52 \pm 4.15$ \\
		\hline
	
		\textbf{Avg. no. of edges} & $142 \pm 3.39$ & $99 \pm 8.47$ \\
		\hline
	
	    \textbf{\# of unique node labels} & 4089 & 64 \\
	    \hline
	
		\textbf{Avg. clustering coeff.} & $0.0009 \pm 0.004$ & $0.0724 \pm 0.026$ \\
		\hline

	\end{tabular}
	
	\label{tab:data_stats}
	
\end{table}

\subsection{Dataset Description}
\label{sect:dataset}
In all our experiments, we use simulated metagenomes using published bovine and its pathogen reference genomes. 
Reference genomes were downloaded from the NCBI nucleotide database. Base-by-base simulation of Illumina (Illumina Inc., San Diego, CA, USA) reads was generated using ART simulator, which simulates user-defined quality score distributions and error rates for the first and second Illumina sequencing reads. Sequence output, generally referred to as reads, was simulated with a minimum Q-score (Quality Score) of 28 and a maximum Q score of 35. ART follows empirical error rate calculations based on the Q-score used for simulation. Following the simulation, a decreasing ratio of the pathogen (Mannheimia haemolytica) genomic reads was added to the 5,000,000 reads of the bovine genome to simulate large metagenome datasets from bovine lung samples.

To make our experiments simple we prepare two randomly sampled datasets with equal distribution of host and pathogen (class variable) from a huge collection of metagenome sequences. We use one small data with \textbf{500 samples} of each class and a large data with \textbf{5000 samples} of each class.

\emph{We make dataset samples used in our experiments to be available for public upon request.}

\subsection{k-\textit{mer} and De-Bruijn Graphs}
\label{sect:kmer}
A metagenome sequence of length $n$ in a genome data can be represented as $X={N_1, N_2, N_3, ... N_n}$, where $N_i$ is a nucleotide and each nucleotide is represented with one of four characters: $N_i \in {A,C,T,G}$. These complex biological sequences are required to be converted into numeric features to perform many machine learning tasks. We use k-\textit{mer} to break each metagenome sequence of length $N$ into $N-k+1$ small sub-sequences of length $k$. A series of these small sub-sequences in each metagenome sequence is often considered as words in sentences and is used to extract numeric features from semantic representations in the sequence. 
Our aim in this work is to efficiently detect pathogen sequences with minimum algorithm run time from large corpus genome data. As we increase the \emph{k} value, the complexity of extracting sub-sequences and extracting feature representations increases as given in the following sections. Thus we experiment with only minimum \emph{k} values in our two data samples. Thus for the small dataset (\emph{DS500}), we use $k=6$ and for the large dataset (\emph{DS5000}), we use $k=3$. However, we give machine learning models performance for multiple \emph{k} in Section~\ref{sect:results}.

\begin{figure}

	\centering
	\includegraphics[scale=0.6]{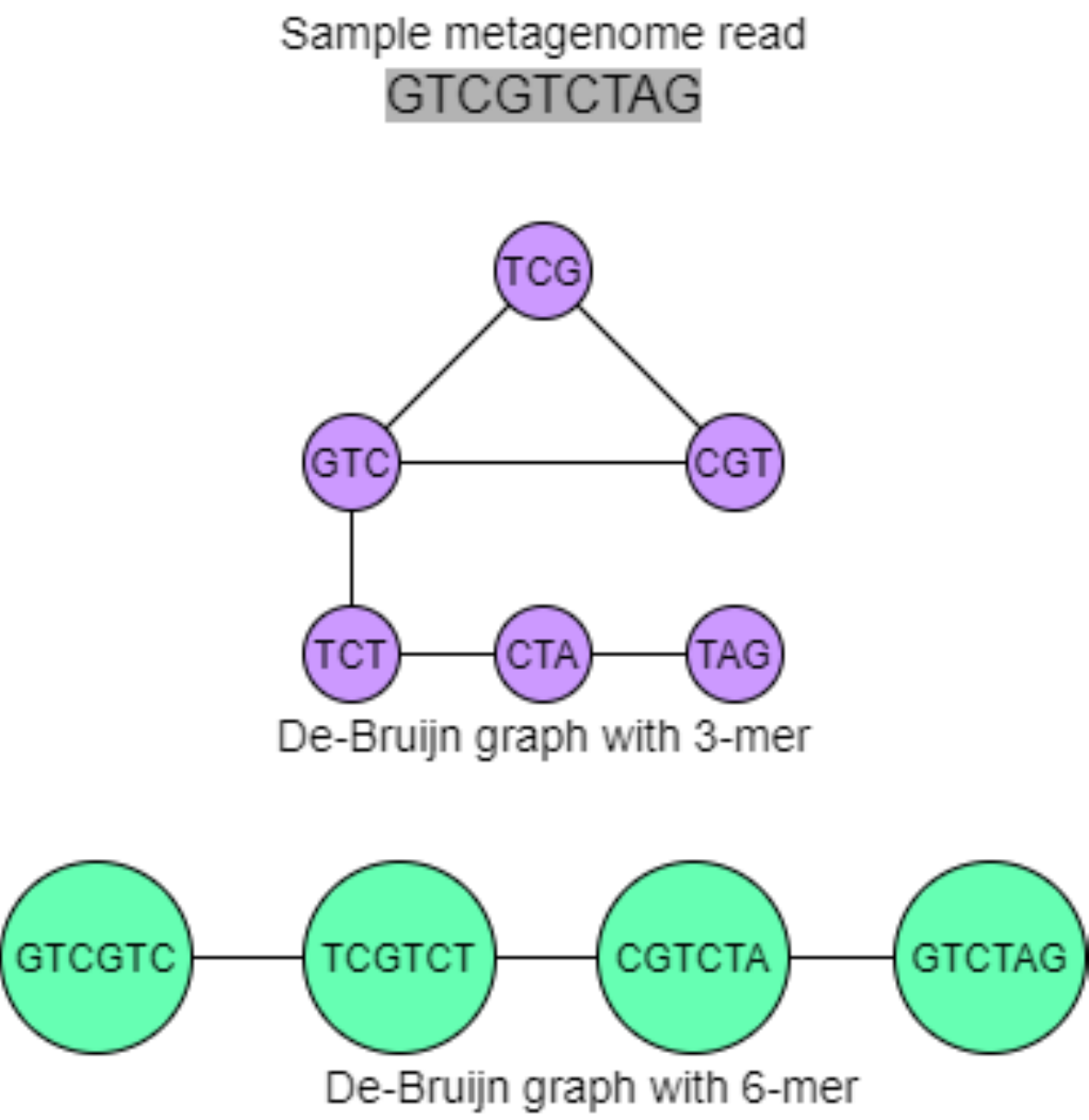}
	\caption{Toy example for mapping a sample metagenome read to De-Bruijn graph with  k-\textit{mer} using \textit{k=(3,6)}. For longer reads the graphs become bigger and more connected.}
	\label{fig:debruijn_toy}
\end{figure}

We use a classic method to assemble the series of k-\textit{mer} sub-sequences in a graph/network structure called \emph{De-Bruijn graphs}. We convert each series of metagenome sub-sequences into a De-Bruijn graph $\mathcal{G} = (\mathcal{V},\mathcal{E})$, where $\mathcal{V}$ is the set of nodes/vertexes, $\mathcal{E}$ is the set of edges, and $|\mathcal{V}| \leq N-k+1$. Each vertex $\mathcal{V_i} \in \mathcal{V}$ represent a k-\textit{mer} and we label each vertex based on the k-\textit{mer}. Also, each edge $\mathcal{E}_i = (\mathcal{V}_i,\mathcal{V}_j)$ represents an ordered sequence of k-\textit{mer} $\mathcal{V}_i$ and $\mathcal{V}_j$ where $\mathcal{V}_i,\mathcal{V}_j \in \mathcal{V}$. A small toy example to convert a very small metagenome sequence into De-Bruijn graphs is given in Figure~\ref{fig:debruijn_toy}. Unlike given in this example the number of nodes in De-Bruijn graphs increase exponentially in real world data when the \emph{k} value increases in k-\textit{mer}. In Table~\ref{tab:data_stats} we give basic statistics of De-Bruijn graphs constructed from our two data samples (\emph{DS500} and \emph{DS5000}). As reported in Table~\ref{tab:data_stats}, the number of nodes in a graph and total number of unique nodes (node labels) increase exponentially as the value of \emph{k} increases, and because of this the amount of clustering varies significantly as given in the clustering coefficient.

\section{Methodology}\label{sec:method}
\subsection{Graph Embedding}\label{sec:embed}
Network/graph embedding can be of two types: mapping vertexes in a graph to a lower-dimensional feature space~\cite{grover2016node2vec,hamilton2017inductive} and mapping the given graph itself into a vector space~\cite{narayanan2017graph2vec,zhang2018end}. In this work, we consider the latter method of graph embedding. Given a set of $M$ graphs $G = {G_1, G_2, ..., G_M}$, where each graph $G_i \in G$ is a set of vertexes and edges: $G_i = (V,E)$, graph embedding is a learning function $\Theta: G_i \rightarrow \mathbb{R}^d$ that translates a graph $G_i$ to a low dimensional feature vector of size $d$. The learning function $\Theta$ uses graph topology, learns the organization of nodes and subgraphs, and embeds the learned information with a statistical or deep learning approach. The learned objective of these feature representations is to cluster together with the topologically similar graphs. In this work, we use a set of $N$ labeled De-Bruijn graphs $\mathcal{G} = {\mathcal{G}_1, \mathcal{G}_2,..., \mathcal{G}_N}$ to learn graph embedding of each De-Bruijn graph $\mathcal{G}_i$. We exploit multiple graph embedding techniques to capture the best feature representation from our De-Bruijn graphs for the following binary classification task. In particular, we use multiple \emph{graph kernel} approaches, and unsupervised network embedding approaches. 

\subsubsection{Graph Kernels} \emph{Graph kernels}~\cite{vishwanathan2010graph} are preliminary forms of obtaining graph features based on graph substructures like shortest paths~\cite{borgwardt2005shortest}, random walks~\cite{kang2012fast}, graphlets~\cite{shervashidze2009efficient}, and isomorphism~\cite{shervashidze2011weisfeiler}. The main objective of graph kernels is to measure the similarity between all graph pairs in the corpus-based on the substructures mentioned above, which result in a feature matrix $\mathbb{R}^{MxM}$ given that there are $M$ graphs in the corpus.

\subsubsection{Unsupervised Representation Learning}
Unlike graph kernels, unsupervised network representation learning tasks learn the feature space directly from given graphs with their structure organization. We use two existing methodologies to obtain network embeddings.

\begin{itemize}
    \item \textbf{node2vec}~\cite{grover2016node2vec} is a local, node-level embedding model which learns low-dimension representations of \textit{each node} in a graph by optimizing a neighborhood preserving objective function. It uses iterations of random walks to gather neighborhood details to learn contextualized representations of nodes that preserves structural equivalence and homophily. For a De-Bruijn graph $\mathcal{G}_i=(\mathcal{V}_i,\mathcal{E}_i)$, we use node2vec embedding to construct local, node-level representations for $v \in \mathcal{V}_i$. We obtain global, graph-level representation of the De-Bruijn graph $\mathcal{G}_i$ by averaging all representations of nodes in $\mathcal{G}_i$.

    \item \textbf{graph2vec}~\cite{narayanan2017graph2vec} is a transductive approach to constructing graph embedding from node labels given by Weisfeiler-Lehman graph kernel~\cite{shervashidze2011weisfeiler} and random walks. graph2vec assigns node labels given by WL graph kernels as initial representation. Similar to node2vec, node representations are averaged with representations from its neighbors. This graph representation is passed to \emph{LSTM} encoder function to get final graph representations.
\end{itemize}

\subsection{Multi-task Deep Networks for Genome Classification}
For classifying each genome sequence, we use a deep, multi-layer neural network. 
The proposed neural architecture has three sub-networks - an encoder, a classification network, and a decoder network. 
The encoder network is used to learn a lower-dimensional feature representation from the $n$-dimension input embedding. This representation is called the latent space and is typically trained to ignore any noise and model the underlying pattern. A decoder network is used to reconstruct the input embedding from the latent space. Finally, the classification network aims to distinguish between the class variables using the latent space. 
This process is represented in Figure \ref{fig:framework}, where the entire framework is outlined. Combined, these networks are trained in a \textit{multi-task} learning setting~\cite{caruana1997multitask}, where the two tasks are classification and input reconstruction, respectively.

\subsubsection{The Need for Multi-task Learning} 
In traditional, feed-forward neural networks used for the classification task, the training objective is to learn internal representations that allow for robust classification of the input. However, given that the genome sequences from the host and pathogen can have highly overlapping k-\textit{mers} and hence similar network embedding, the encoder can over-fit to the training distribution due to the highly specific objective function. 
To overcome this limitation, we propose the use of a decoder network \textit{and} a classification network trained in tandem with the encoder network. The objective function function now becomes the learning of an internal representation that jointly models the underlying distribution for both classification and input reconstruction. The encoder network and the latent space are \textit{shared} between the classification and reconstruction heads and hence reduces the risk of over-fitting by providing implicit regularization and reducing the representation bias in the network. Here, representation bias refers to the tendency of neural networks to learn representations that are highly specific to a certain task and associated training data that prevent the model from generalizing to unseen samples.

The multi-task loss function for the proposed framework is given by
\begin{equation}
    \mathcal{L}_{total} = \lambda_1\mathcal{L}_{ce} + \lambda_2\mathcal{L}_{rc}
    \label{eqn:loss}
\end{equation}
where $\mathcal{L}_{CE}$ refers to the weighted cross entropy loss and $\mathcal{L}_{RC}$ refers to the reconstruction loss from the decoder head; $\lambda_1$ and $\lambda_2$ are modulating factors to balance the loss function between classification performance and reconstruction penalty. The reconstruction loss is an L2 difference between the reconstructed embedding ($\hat{x}$) and the actual input embedding ($x$) and is given by $\sum_{i=1}^{n} {\lVert x - \hat{x} \lVert_{\ell_2}^2}$, where $n$ is the dimension of the input embedding. 

Note that this is different from only pre-training the encoder and decoder networks as an \textit{auto-encoder}. While the auto-encoder objective is to reconstruct the original input through a compressed representation, there can exist an inherent representation bias which causes the network to produce low quality reconstructions and hence a poor latent space representation of unseen training examples. The low \textit{inter-class} variation in genome sequences can cause the network to learn noisy representations and hence reduce the classification accuracy. We highlight the importance of the multi-task learning objective function in Section~\ref{sect:results}, with a neural network baseline that is trained only with the classification loss after pre-training as an autoencoder with the L-2 reconstruction loss.

\subsubsection{Implementation Details}
Since the proposed architecture has a complex structure, we provide the implementation details here. 
The encoder network has five (5) fully connected (dense) layers. We intersperse each encoder layer with a dropout layer~\cite{srivastava2014dropout}, with a dropout probability of $50\%$. We reduce the dimensions of the input by $0.5\times$ at each fully connected (dense) layer. 
The decoder network (for reconstruction) consists of two densely connected layers to increase the encoded features back to the original dimension. 
The classification network has two (2) densely connected layers that take the encoded representation as input and produces the genome classification. This is the only part of the network that is trained in a supervised manner. The encoder and decoder networks are trained in an unsupervised manner.

Due to the limited training data and the low \textit{inter-class} variation, neural networks can over-fit to the training data and not generalize to any variations induced by noise in the observation. To overcome this, we propose the following training protocol. 
First, we perform a \textit{cold start}, i.e. for ten epochs, we train the encoder and decoder networks in as a traditional auto-encoder with a very low learning rate ($1\times10^{-8}$). Then, we freeze the decoder network and train the encoder and classifier branch in a supervised manner for $10$ epochs with a learning rate of $4\times10^{-4}$. Finally, the entire network is train end-to-end for $15$ epochs with a learning rate of $4\times10^{-5}$. The learning rate schedule and the varying objective functions help prevent over-fitting and learn robust features for classification. 

\section{Experiments and Results}
\label{sect:results}
\begin{table*}[h!]
	\centering
	
	\caption{Classification accuracy of multiple graph embedding techniques with SVM and Deep Learning (DL) on graphs constructed with 6-mer from 1000 samples (500 - positive and 500 - negative)}
	\begin{tabular}{P{11em}|P{6em}|P{6em}|P{6em}|P{6em}}
		\hline

        \textbf{Baseline} & \multicolumn{4}{c}{47.29} \\
        \hline
        \hline

		\textbf{Embedding Type} & \textbf{LG} & \textbf{SVM} & \textbf{NN} & \textbf{DL} \\
		\hline
	
		\textbf{SPK} & 53.58 $\pm$ 0.05 & 53.1 & 58.59 $\pm$ 0.23  & 63.80 $\pm$ 0.19\\
		\hline
	
		\textbf{WLK} & 55.87 $\pm$ 0.06 & 53.8  & 57.81 $\pm$ 0.53  & 61.41 $\pm$ 0.38\\
		\hline
	
		\textbf{GSK} & 58.33 $\pm$ 0.07 & 52.9  & 56.25 $\pm$ 0.51 & 60.16 $\pm$ 0.95\\
		\hline
	
		\textbf{RWK} & 59.34 $\pm$ 0.11 & 50.8  & 54.69 $\pm$ 0.83  & 56.41 $\pm$ 0.72\\
		\hline
	
		\textbf{node2vec} & 56.53 $\pm$ 0.23 & 56.8 $\pm$ 0.11  & 63.28 $\pm$ 0.96 & \textbf{73.49 $\pm$ 0.69}\\
		\hline
	
		\textbf{graph2vec} & 58 & 52.16  & 60.19 & 66.42 \\
		\hline

	\end{tabular}
	
	\label{tab:acc1}
	
\end{table*}

In this section we give evaluation of multiple graph embedding techniques by comparing the proposed deep learning classifier with multiple classification algorithms in pathogen prediction from datasets described in Section~\ref{sect:dataset}. Along with model performance we also report results obtained by tuning parameters in de-bruijn graph generation and graph embedding algorithms.

\subsection{Baseline Models}\label{sec:baselines}
We conduct the following baseline experiments to extract graph embeddings in our study to compare developed models based on network embedding and deep learning classifier.

\subsubsection{Naive and unsupervised approach}
In this simple approach, we utilize k-mers extracted as described in Section~\ref{sect:kmer} from metagenome sequences itself with a simple unsupervised KMeans ($k=2$) algorithm to cluster host and pathogen metagenome sequences. For this approach we first generate k-mer sub-sequences ($k=6$ for \emph{DS500} and $k=3$ for \emph{DS5000}). For each metagenome sequence $\mathcal{M}$ of length $N$ we generate a feature vector of shape $1 \times \mathcal{P}$, where $\mathcal{P} = (N-k+1)$ and each component in the feature vector represents the normalized frequency of the presence of the defined k-mer subsequence. With this feature vector, we use KMeans algorithm to separate host and pathogen metagenome sequences. 

\subsubsection{Graph kernels}
We use the following 4 types of graph kernels to compute graph similarities with other graphs. 

\begin{itemize}
    \item \textbf{Shortest Path kernel (SPK)}~\cite{borgwardt2005shortest}: Similarity is determined by comparing lengths of shortest paths between nodes in two graphs
    \item \textbf{Weisfeiler-Lehman kernel (WLK)}~\cite{shervashidze2011weisfeiler}: Graph similarity is calculate based on graph isomorphism
    \item \textbf{Graphlets Sampling kernel (GSK)}~\cite{shervashidze2009efficient}: Based on the number of same subgraph structures - \emph{graphlets} - in two graphs
    \item \textbf{Random Walk kernel (RWK)}~\cite{kang2012fast}: It determines similarity of two graphs with the number of matching set of random walks on two graphs
\end{itemize}

Each graph kernel generate $N \times N$ feature space matrix.

\subsubsection{Unsupervised Graph Representations}
For both \emph{node2vec} and \emph{graph2vec} we use the output embedding size as $128$, number of walks per node as $10$, and walk length as $80$. For node2vec we use parameters $p=q=1$ and $window\_size=10$. For graph2vec, we use parameters $height=3$.

\subsection{Experiment Setup}
Along with the performance of our deep learning classifier, we use the following vector space models with the given parameters to compare the classification performance. We use 10-fold cross validation result for all datasets and embedding types. We run 10 iterations of each cross validation test and report the average accuracy and standard deviation.

\subsubsection{Logistic Regression}
We use simple Logistic Regression with \emph{C}$=10$, \emph{solver} as \emph{lbfgs}, and \emph{penalty} as \emph{l2}.

\subsubsection{Support Vector Machine}
We use C-SVM classifier from the LIBSVM~\cite{chang2011libsvm}. We used grid search and cross validation to determine the value of \emph{C} and \emph{kernel type} in the classifier. For all datasets and embeddings, we use $C=1.0$ and kernel type as \textbf{radial bias function}.

\subsubsection{Neural networks}
We evaluate the proposed model (described Section~\ref{sec:method} and a standard neural network baseline. We denote these models as \textit{DL} and \textit{NN}, respectively. The \textit{NN} baseline has the same network structure as the \textit{DL}, but does not have the mult-task training objective. It is trained with the weighted cross-entropy loss function for classification, after pre-training for $10$ epochs as an auto-encoder. 

\subsection{Hyperparameter Selection}
For configuring the learning rate for the \textit{DL} and \textit{NN} models, we performed a sensitivity analysis of the learning rate using a grid search. We used a grid search on the learning rate using a log scale from $0.1$ to $10^-6$ to highlight the order at which the best learning rate could be found. The process was repeated the two datasets since they had different batch sizes - $32$ (for the smaller dataset) and $1024$ (for the larger dataset). The dropout probability was set to be $0.5$ after manual tuning between values of $0.25$ and $0.75$. The modulating factors $\lambda_1$ and $\lambda_2$ (from Equation~\ref{eqn:loss}) were set to be $2$ and $0.5$, respectively after manual tuning, with the range of values tried being $0.5$ and $2$, in increments of $0.5$. We find that the use of $\lambda_1$ and $\lambda_2$ to modulate the final loss allows for faster convergence during training. 

\begin{table*}[h!]
	\centering
	
	\caption{Classification accuracy of multiple graph embedding techniques with SVM and Deep Learning (DL) on graphs constructed with 3-mer from 10000 samples (5000 - positive and 5000 - negative)}
	\begin{tabular}{P{11em}|P{6em}|P{6em}|P{6em}|P{6em}}
		\hline

        \textbf{Baseline} & \multicolumn{4}{c}{32.14} \\
        \hline
        \hline

		\textbf{Embedding Type} & \textbf{LG} & \textbf{SVM} & \textbf{NN} & \textbf{DL} \\
		\hline
	
		\textbf{SPK} & 60.25 $\pm$ 0.14 & 61.27 & 57.23 $\pm$ 0.37 & 67.9 $\pm$ 0.29\\
		\hline
	
		\textbf{WLK} & 62.4 $\pm$ 0.09 & 61.11 & 57.03 $\pm$ 0.28 & 61.6 $\pm$ 0.89\\
		\hline
	
		\textbf{GSK} & 64.76 $\pm$ 0.12 & 64.83 & 60.93 $\pm$ 0.16 & 62.81 $\pm$ 0.76\\
		\hline
	
		\textbf{RWK} & - & - & - & -\\
		\hline
	
		\textbf{node2vec} & 78.84 $\pm$ 0.14 & 83.7 & 81.78 $\pm$ 0.62 & \textbf{89.74 $\pm$ 0.60} \\
		\hline
	
		\textbf{graph2vec} & 57.45 $\pm$ 0.01 & 54.15 & 55.83 & 59.62\\
		\hline

	\end{tabular}
	\label{tab:acc2}
	
\end{table*}

\subsection{Quantitative Evaluation}
In Tables~\ref{tab:acc1} and \ref{tab:acc2} we give average accuracy (in percentage) and standard deviation for \emph{DS500} and \emph{DS5000} respectively. We report the standard deviation only when it is significant i.e. when the observed standard deviation is greater than $0.01\%$. 
We report results of each graph embedding model (SPK, WLK, RWK, GSK, node2vec,and graph2vec) with each classifier (LG, SVM, NN, and DL) as discussed in the previous section from 10 iterations of 10-fold cross validation.
Throughout our experiments, we compare the results of our small data sample (\emph{DS500}) and large data sample (\emph{DS5000}). 

We outperform the na\"{i}ve k-\textit{mer} frequency-based feature representation by a large margin on both datasets. 
It can be seen that the proposed deep neural network with the multi-task training objective outperforms the other baselines, including a similar neural network \textit{without} the multi-task objective. The competitive performance of the logistic regression and SVM models indicate that the learned embedding (constructed from the De-Bruijin graphs) are a good feature representation of the genome sequences.

Among the various embedding approaches, node2vec outperforms all other approaches by a large margin on both datasets, providing gains of $26.2\%$ over the na\"{i}ve k-\textit{mer} frequency features and $7.02\%$ over the closely related graph2vec embedding, on the DS500 dataset. Similar gains can be seen on the DS5000 dataset. Among graph kernels, the shortest path graph kernel (SPK) embedding provides a competitive performance and the graphlets sampling kernel (GSK) is more resilient across the different classifier approaches.

\subsection{Ablative Studies}
We also evaluate different variations of our approach to test the effectiveness of each component in the framework. Namely, we evaluate two variations: (i) the effect of the length of sub-sequence, and (ii) the effect of multi-task learning.

\subsubsection{Length of Sub-sequence (\textit{k})}
\begin{figure*}
    \centering
    \includegraphics[scale=0.99]{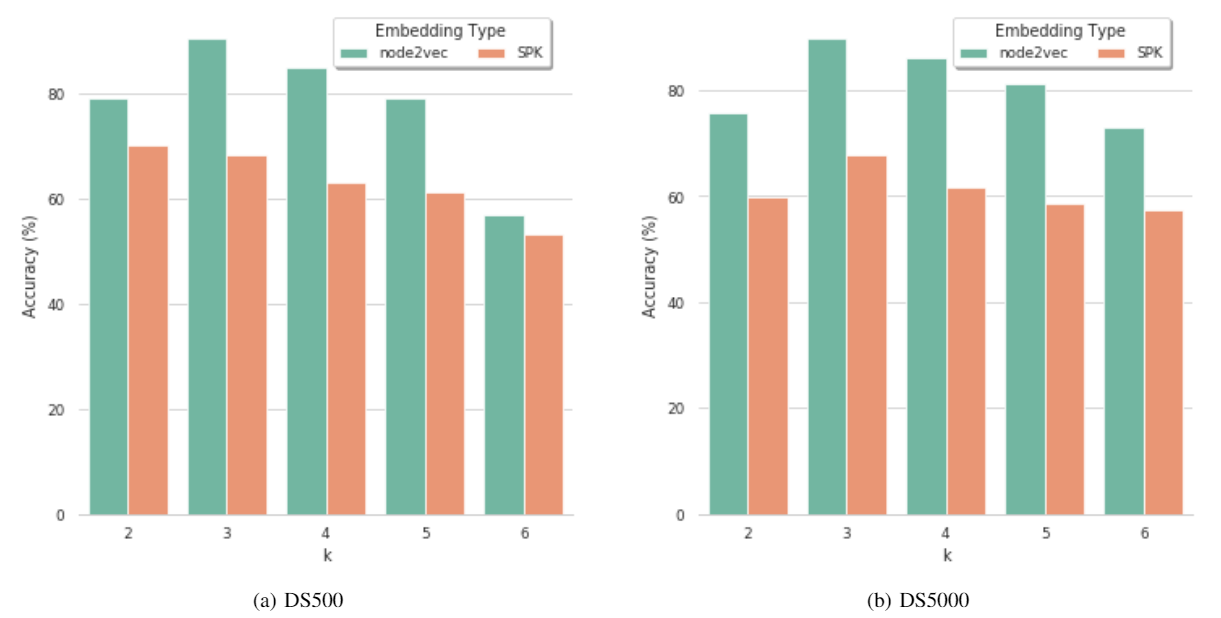}
    \caption{Accuracy of deep learning classification for graph embedding with shortest path graph kernel and node2vec with $k={2,3,4,5,6}$ for (a) DS500 and (b) DS5000 datasets}
    \label{fig:k_accuracy}
\end{figure*}

First, we vary the sub-sequence length $k$ used to construct the De-Bruijin graphs and evaluate the performance of our full model (DL) on both the DS500 and DS5000 datasets. 
We use two of the best performing embedding models (node2vec and shortest path graph kernel) and summarize the results in Figure~\ref{fig:k_accuracy}. 
We find that node2vec outperforms SPK at all values of $k$. 
Interestingly, increasing $k$ has a consistently detrimental effect on both embedding methods across the two datasets. 
This could arguably be attributed to the fact that the number of unique nodes increases with increase in $k$ and hence the amount of variability is more pronounced. The node2vec embedding is the most affected with the change in the sub-sequence length, especially for the DS500 dataset, with the accuracy dropping by almost $17\%$. The SPK embedding provides more consistent performance across the sub-sequence lengths. This can be attributed to the presence of less \textit{tottering walks}~\cite{borgwardt2005shortest} in the De-Bruijin graphs that are constructed from the genome sequence. 

\subsubsection{Effect of Training Objective}
We also perform ablations on the proposed deep learning model. Specifically, we keep the network structure static and vary the training objective to not include the multi-task learning setting. Instead, we train the network in a single-task, classification setting, as is the case with traditional neural network applications. As can be seen from both Table~\ref{tab:acc1} and Table~\ref{tab:acc2}, the use of the multi-task objective function provides consistent improvements on both datasets and across most embedding methods. On average, the multi-task objective provides an average improvement of $6.65\%$ on the DS500 dataset and $5.77\%$ on the DS5000 dataset, indicating the tendency to prevent over-fitting on smaller datasets, for which it was designed. Additionally, we find that training the network \textit{without a decoder network}, reduces the accuracy of the model by an average of $8.43\%$ on the DS500 dataset and by $7.64\%$ on the DS5000 dataset. 

%
%

\section{Conclusion and Future work}
Detection of pathogens like \emph{Mannheimia Hemolytica} is of huge importance in animal diagnostics, particularly with BRDC. In this work we proved that machine learning models equipped with network embedding and deep learning classifier can help identifying pathogens metagenome sequences from host metagenome sequences. With our experiments we showed that the adapted machine learning approach help predicting  pathogen metagenome sequences with a huge margin in accuracy from baseline models. 

This work is conducted completely on small and large data samples from simulated genomic data to validate the role of machine learning in veterinary medicine. For simplicity we generated the simulated genome sequences to contain only single pathogen in the host and also generated balanced sequences of host and pathogen. In the real world data, this scenario is quite the opposite. The real world animal genome sequences have limited pathogen sequences and the data is completely unbalanced. Given the better performance from this work, there are multiple ways to continue this work in the future. An end-to-end framework specifically for pathogen detection in animal genome sequences can be developed. This framework can utilize ensemble learning to learn from multiple learning models to obtain best prediction accuracy. Importantly, more sophisticated models that learn from unbalanced data distribution to predict multiple pathogen sequences can be proposed. Also pathogen sequences are relative to host. For example, the pathogen used in this work - \emph{Mannheimia Hemolytica} - is not necessarily a pathogen for all other hosts. This relative mapping between host and pathogen can be considered for future prediction models.

\section*{Acknowledgment}

This research was supported in part by the US Department of Agriculture (USDA) grants AP20VSD and B000C011.

\bibliographystyle{IEEEtran}
\bibliography{bibliography}

\end{document}